\pdfoutput=1
\documentclass[10pt,logo,copyright]{nv}

\usepackage{wrapfig}
\usepackage{algorithm}
\usepackage{algorithmic}
\usepackage{subcaption}
\usepackage{mathtools}
\usepackage{dsfont}
\usepackage{multirow}
\usepackage{makecell}
\usepackage{adjustbox}
\usepackage{diagbox}
\usepackage{bbding}
\usepackage{pifont}
\usepackage{xspace}
\IfFileExists{tcolorbox.sty}{%
  \usepackage{tcolorbox}
  \tcbuselibrary{listings,breakable,skins}
}{}
\usepackage{listings}
\usepackage{natbib}
\usepackage{soul}  % \hl{} for line-breakable value-token highlighting

\definecolor{symbg}{RGB}{255,220,220}   % 默认红背景
\definecolor{valbg}{RGB}{220,230,250}   % value 背景
\definecolor{boxbg}{RGB}{248,248,248}   % 整体背景

\newcommand{\val}[1]{\colorbox{valbg}{\texttt{#1}}}

\definecolor{symcolor}{RGB}{150,30,30}   % 其他字符
\definecolor{valcolor}{RGB}{20,90,160}   % value内容
\definecolor{nvidiagreen}{HTML}{76B900}

\definecolor{darkred}{rgb}{0.459,0.0,0.08}
\definecolor{myblue}{HTML}{47B1E1}
\definecolor{mygreen}{HTML}{46D45A}
\definecolor{myorange}{HTML}{F3AA84}
\definecolor{usercolor}{HTML}{B41601}
\definecolor{assistantcolor}{HTML}{76B900}
\definecolor{linkc}{rgb}{0,0.44,0.74}
\definecolor{eqc}{rgb}{1,0,0}
\definecolor{newcitecolor}{rgb}{0,0.6,0}

\colorlet{valbg}{nvidiagreen!22}
\sethlcolor{valbg}
\renewcommand{\val}[1]{{\ttfamily\hl{#1}}}

\newcommand{\modelname}{Fast-dDrive\xspace}

\lstdefinestyle{jsonstyle}{
  basicstyle=\ttfamily\scriptsize,
  showstringspaces=false,
  breaklines=true,
  breakatwhitespace=false,
  frame=none,
  backgroundcolor=,
  escapeinside={|}{|},
}
\title{\modelname: Efficient Block-Diffusion VLM for Autonomous Driving}

\author{
\textbf{Kewei Zhang}\textsuperscript{1*} ~~~
\textbf{Jin Wang}\textsuperscript{3,2*} ~~~
\textbf{Sensen Gao}\textsuperscript{2} ~~~
\textbf{Chengyue Wu}\textsuperscript{3,2} ~~~
\textbf{Yulong Cao}\textsuperscript{2} \\
\textbf{Songyang Han}\textsuperscript{2} ~~
\textbf{Boris Ivanovic}\textsuperscript{2} ~~
\textbf{Langechuan Liu}\textsuperscript{2} ~~
\textbf{Marco Pavone}\textsuperscript{2} ~~
\textbf{Song Han}\textsuperscript{4,2} \\
\textbf{Daquan Zhou}\textsuperscript{1\dag} ~~
\textbf{Enze Xie}\textsuperscript{2\dag} \\
\vspace{2mm}
{\normalsize
\textsuperscript{1}Peking University ~~
\textsuperscript{2}NVIDIA ~~
\textsuperscript{3}The University of Hong Kong ~~
\textsuperscript{4}MIT \\
\textsuperscript{*}Equal contribution \quad \textsuperscript{\dag}Co-lead
}
}

\begin{abstract}
End-to-end autonomous driving via Vision-Language-Action (VLA) models demands a precarious balance between high-fidelity trajectory planning and efficient inference. Existing paradigms typically fall short: autoregressive (AR) VLAs are memory-bandwidth-bound on edge hardware and prone to exposure-bias drift, while full-sequence diffusion models preclude KV-cache reuse and suffer from ``logical leakage'' that violates the fundamental perceive-then-plan causality. We present \modelname, a block-diffusion VLA that performs bidirectional refinement within semantic units while enforcing strict causal ordering across them. Leveraging the observation that driving VLAs often emit structured JSON-like outputs, \modelname freezes structural tokens into a section scaffold 
and employs a section-aware training recipe that prioritizes safety-critical planning. We further introduce Scaffold Speculative Decoding to achieve AR-equivalent quality at significantly higher throughput. 
Finally, we propose a low-overhead test-time scaling scheme: by forking $N$ stochastic trajectory rollouts from a single shared-prefix KV cache and averaging them, we effectively suppress prediction variance at a fractional computational cost. Empirical results demonstrate that \modelname redefines the speed-accuracy frontier for driving agents. On the WOD-E2E test set, \modelname achieves SOTA ADE@3s and ADE@5s, alongside the highest RFS among diffusion-based VLAs; on nuScenes, it reduces average L2 error to $0.32$m (a $22\%$ improvement). When integrated with SGLang, our framework delivers $12\times$ throughput speedup over the AR baseline, narrowing the gap between high-capacity VLAs and the efficiency demands of real-time on-vehicle deployment.

\textbf{Links:}\quad ~\href{https://github.com/NVlabs/Fast-dLLM}{\color{nvidiagreen}Github Code}~$|$~\href{https://nvlabs.github.io/Fast-dLLM/fast_ddrive/}{\color{nvidiagreen}Project Page}
\end{abstract}

\begin{document}

\maketitle
\section{Introduction}

End-to-end (E2E) autonomous driving has progressed rapidly by unifying perception, reasoning, and planning within a single trainable system~\citep{hu2023planning,jiang2023vad,xu2025wod}. A growing line of work extends this paradigm with Vision-Language Models (VLMs) and Vision-Language-Action (VLA) models~\citep{tian2024drivevlm,zhouautovla,rowe2025poutine,ma2025dvlm}, which leverage broad world knowledge and natural-language reasoning to handle the long-tail scenarios that dominate real-world driving and to expose interpretable explanations of the agent's decisions. For any such system to be practically useful, two requirements must be met \emph{simultaneously}: the predicted trajectory must be accurate and globally consistent with the model's reasoning, and inference must be efficient enough on edge hardware at batch size one to remain competitive with classical planners. Existing VLAs typically satisfy at most one of these criteria.

\begin{figure}[t]
    \centering
\includegraphics[width=\linewidth]{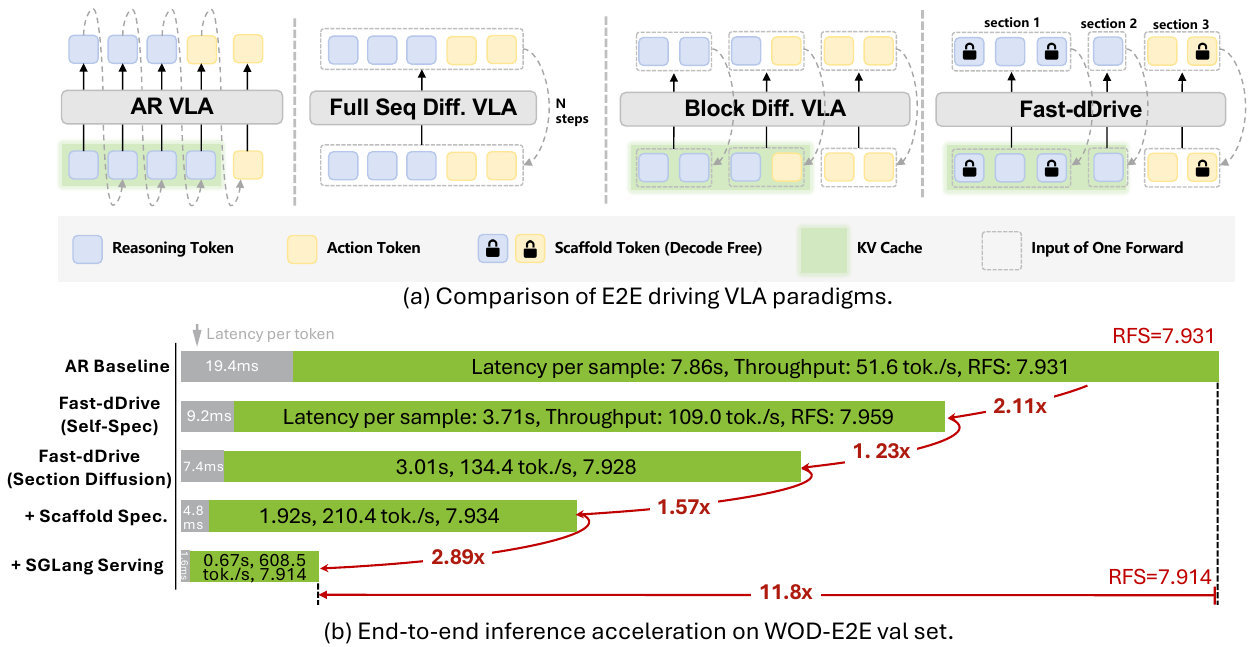}
    \caption{(a) Comparison of E2E driving VLA paradigms. AR VLAs are memory-bandwidth-bound at batch size one and produce only one token per forward pass; full-sequence diffusion VLAs preclude KV-cache reuse and admit logical leakage across the perceive-then-plan stages. \modelname overcomes both via section-aligned block diffusion, and further pre-fills template tokens as a frozen scaffold to accelerate inference and inject schema priors. (b) Our combined approach achieves up to 11.8$\times$ end-to-end speedup compared to the AR baseline, measured on a single NVIDIA H100 GPU.}
    \label{fig:overview}
\vspace{-20pt}
\end{figure}

Driving VLAs are predominantly built on autoregressive (AR) decoders inherited from general-purpose VLMs~\citep{liu2023visual,bai2025qwen2}, which emit the structured reasoning trace and the trajectory tokens one at a time. Sequential decoding causes a well-known \emph{exposure-bias} effect: each waypoint conditions on previously emitted (and possibly noisy) coordinates, so small errors at the start of a 5\,s plan can compound into physically implausible maneuvers~\citep{huang2025survey}. In addition, single-token decoding at batch size one is strictly memory-bandwidth-bound on modern GPUs: each new token reloads the full set of model weights while leaving the available parallel compute largely idle, making efficient on-vehicle deployment fundamentally hard~\citep{wu2025fastdvlm,wu2025fast_dllm}.

Recent diffusion-based language models~\citep{nie2025large,you2025llada,yu2025dimple}, typically formulated as masked-diffusion modeling (MDM) where masked tokens are iteratively unmasked via bidirectional attention, replace AR with iterative denoising that provides global context at every refinement step. Applied to driving, dVLM-AD~\citep{ma2025dvlm} reformulates the structured driving response as a single bidirectional denoising target and improves reasoning--action consistency over AR baselines, but at two structural costs: (i) full-sequence bidirectional attention precludes KV-cache reuse, keeping end-to-end latency far above AR baselines; and (ii) treating the response as one bidirectional unit ignores its inherent causal structure (perception, explanation, meta-behavior decision, and trajectory in that order), admitting \emph{logical leakage} where the planned trajectory can retroactively influence the model's stated perception. We instead propose \modelname (Figure~\ref{fig:overview}), a block-diffusion VLA that decodes the structured driving output section by section under strict causal ordering, with bidirectional refinement confined within each section, directly resolving both costs while preserving the global-context benefit of diffusion.

On top of this paradigm, \modelname{} further exploits a structural observation about modern driving VLMs~\citep{ma2025dvlm,rowe2025poutine,zhouautovla}: their structured outputs bundle perception, chain-of-thought, and trajectory into a schema-defined JSON whose keys and syntax are determined entirely by the schema rather than by the model~\citep{gu2026structuredcot}. We treat those deterministic tokens as a frozen \emph{scaffold} and denoise only the value tokens, concentrating model capacity on the few positions that actually require prediction. Building on this scaffold and the Fast-dVLM~\citep{wu2025fastdvlm} architecture with a Qwen2.5-VL-3B~\citep{bai2025qwen2} backbone, our contributions span three axes: a section-weighted, noise-adaptive training scheme that prioritizes safety-critical reasoning; a scaffold-aware self-speculative decoder that auto-accepts structural tokens and verifies an MDM draft with the AR head, delivering AR-quality outputs at substantially lower latency; and a low-overhead test-time inference scaling scheme that, with the deterministic prefix decoded once, samples the AR verifier of Scaffold Speculative Decoding only on the trajectory section and averages a small number of trajectory rollouts forked from a shared KV cache, trading a fraction of additional inference compute for a meaningful accuracy gain. Concretely:

\begin{itemize}[leftmargin=*,nosep,itemsep=2pt]
\item \textbf{Section-Aware Structured Diffusion (SASD).} A scaffold-based training scheme that aligns block boundaries with semantic sections (ensuring $100\%$ structural validity by construction) and uses section-weighted cross-entropy together with a section-adaptive Beta noise schedule to concentrate capacity on safety-critical sections, at zero inference overhead.

\item \textbf{Scaffold Speculative Decoding and shared-prefix test-time scaling.} Scaffold Speculative Decoding (SS) auto-accepts scaffold tokens and lets the AR head verify a parallel MDM draft, producing outputs identical to pure AR at substantially lower latency. We further turn the deterministic SS verifier into a tunable inference-scaling axis: with the prefix decoded once and the verifier sampled at non-zero temperature only on the trajectory section, $N$ trajectory rollouts are forked from a shared KV cache and averaged, trading a fraction of extra inference compute for a meaningful accuracy gain.

\item \textbf{State-of-the-art accuracy at $\mathbf{12\times}$ throughput.} On the WOD-E2E test set, \modelname{} achieves the lowest ADE@3s and ADE@5s among compared methods while maintaining the highest RFS among diffusion-based VLAs. It delivers this SOTA accuracy at over $200$ tokens per second on a single H100—representing a $6\times$ throughput increase over full-sequence diffusion and $4\times$ over AR baselines. When integrated with SGLang, this efficiency gain scales to a $12\times$ speedup over AR baselines, demonstrating that high-capacity VLAs can effectively bridge the gap toward real-time on-vehicle deployment without accuracy compromises.
\end{itemize}

These results indicate block-diffusion VLAs, when paired with structure-aware training and inference, can match or exceed the accuracy of strong AR and full-sequence-diffusion baselines while running at substantially higher throughput, without sacrificing the interpretability of structured CoT outputs.

% \item \textbf{State-of-the-art accuracy at $\mathbf{6\times}$ throughput.} On the WOD-E2E test set, \modelname{} attains the lowest ADE@3s and ADE@5s among compared methods at over $200$ tokens per second on a single H100 ($6\times$ the throughput of full-sequence diffusion and $4\times$ that of AR baselines) while matching the highest RFS among diffusion-based VLAs. On nuScenes val, it lowers the average L2 to $0.32$\,m ($-22\%$ vs.\ dVLM-AD).

\section{Related Work}

\textbf{Vision-Language-Action Models for Autonomous Driving.} Vision-Language-Action (VLA) models unify perception, reasoning, and planning within a single multimodal framework. Autoregressive VLAs 
% such as OpenDriveVLA~\citep{zhou2025opendrivevla}, AutoVLA~\citep{zhouautovla}, and Poutine~\citep{rowe2025poutine} 
leverage language-model reasoning to improve trajectory prediction in long-tail scenarios, with recent works further incorporating chain-of-thought reasoning~\citep{wang2024drivecot,tian2024drivevlm,zhouautovla}. 
% and reinforcement learning fine-tuning~\citep{yuan2025autodrive,luo2025adathinkdrive,rowe2025poutine}
% . \songyang{Do we want to remove the reinforcment learning part if we don't have RL yet. If not, they are not relevant to this work.} 
However, AR decoding is inherently sequential and memory-bandwidth-bound at batch size 1~\citep{wu2025fastdvlm}, a critical efficiency bottleneck for latency-sensitive driving deployments, and the autoregressive factorization introduces exposure bias that compounds waypoint errors over longer horizons.
To address these issues, diffusion-based VLAs have been explored for driving. dVLM-AD~\citep{ma2025dvlm} applies discrete masked diffusion to jointly generate structured reasoning and trajectories, improving behavior-trajectory consistency. Concurrent works~\citep{li2025discrete,wen2025dvla} also adopt discrete diffusion for driving VLAs. However, these methods rely on full-sequence bidirectional diffusion, which precludes KV-cache reuse and incurs high computational overhead. Our work addresses this efficiency gap by adopting block diffusion, enabling parallel generation within blocks while maintaining causal ordering across blocks.

\textbf{Diffusion Large Language Models.} Discrete diffusion for text has progressed from foundational formulations~\citep{austin2021structured,li2022diffusion} through refined masked diffusion objectives~\citep{lou2024discrete,sahoo2024simple,shi2024simplified} to large-scale models such as LLaDA~\citep{nie2025large} and Dream~\citep{ye2025dream} that match autoregressive performance. Post-training methods~\citep{zhu2025llada15variancereducedpreference,wang2025revolutionizingreinforcementlearningframework} further align diffusion LMs with human preferences, and multimodal extensions~\citep{yang2025mmada,you2025llada,yu2025dimple} integrate visual instruction tuning.
A key limitation of full-sequence diffusion LMs is the inability to leverage KV caching. Block Diffusion~\citep{arriola2025block} addresses this by partitioning the output into fixed-size blocks with bidirectional attention \emph{within} blocks and causal attention \emph{across} blocks, recovering KV-cache compatibility. Fast-dVLM~\citep{wu2025fastdvlm} extends this to vision-language models, achieving a significant speedup over AR baselines via direct AR-to-diffusion conversion and self-speculative decoding~\citep{wu2025fast_dllm}. Our work builds upon Fast-dVLM and introduces structure-aware scaffold diffusion with safety-prioritized training that exploits the structured output format of autonomous driving.

\textbf{Efficient Decoding and Test-Time Scaling}. Speculative decoding~\citep{leviathan2023fast,chen2023accelerating} accelerates AR generation by drafting multiple tokens for parallel verification. Self-speculative variants~\citep{zhang2024draft} eliminate the separate draft model by reusing the same model for both drafting and verification. Fast-dLLM~\citep{wu2025fast_dllm} extends this to block diffusion, where the MDM head drafts tokens via bidirectional attention and an AR pass with causal attention verifies the draft. Medusa~\citep{cai2024medusa} and EAGLE~\citep{li2024eagle} propose lightweight draft heads for tree-structured verification, further improving acceptance rates. Our Scaffold Speculative Decoding builds on the self-speculative framework of Fast-dLLM but exploits the known output structure to auto-accept scaffold tokens and skip redundant verification.
Test-time compute scaling has been explored through Best-of-N sampling~\citep{cobbe2021training,lightman2024lets}, reward-guided search~\citep{snell2024scaling}, and multi-modal trajectory selection in diffusion planners~\citep{liao2025diffusiondrive,yang2024diffusion_es}. These approaches typically require a separate verifier or a large sample budget. Our shared-prefix rollout scheme instead exploits the deterministic structure of the first three sections to amortize prefix computation, applying stochasticity only to the trajectory section at a fractional per-rollout cost.

\section{Methodology} \label{sec:method}

We present \modelname, a block-diffusion VLA for end-to-end autonomous driving. We first review the block-diffusion formulation (\S\ref{sec:prelim}), then describe our structure-aware scaffold diffusion training (\S\ref{sec:scaffold}), the two inference modes it admits (Section Diffusion and Scaffold Speculative Decoding, \S\ref{sec:scaffold_spec}), and a low-overhead test-time inference scaling scheme that decodes the deterministic prefix once and averages multiple stochastic trajectory-section rollouts forked from a shared KV cache (\S\ref{sec:merge}). \looseness=-1

\subsection{Preliminaries: Block-Causal Masked Diffusion} \label{sec:prelim}

\paragraph{Masked Diffusion Language Models.}
Let $\mathbf{x}_0 = (x_1, \ldots, x_L)$ be the target token sequence and $\mathbf{c} = (\mathbf{v}, \mathbf{p})$ the conditioning context (visual features and text prompt). A masked diffusion model~\citep{sahoo2024simple} defines a forward process that randomly replaces tokens with a special $[\mathrm{MASK}]$ token according to a noise schedule $\{\lambda_t\}_{t=1}^{T}$, yielding a corrupted sequence $\mathbf{x}_t$. The reverse process applies a denoising policy $p_\theta$ that predicts replacements for masked positions while keeping visible tokens fixed. Training minimizes the masked cross-entropy loss:
\begin{equation}
\label{eq:mdm_loss}
\mathcal{L}_{\mathrm{MDM}}(\theta) = \mathbb{E}_{t, \mathbf{x}_0, \mathbf{x}_t} \left[ -\frac{1}{|\mathcal{M}_t|} \sum_{i \in \mathcal{M}_t} \log p_\theta(x_0^i \mid \mathbf{x}_t, \mathbf{c}) \right],
\end{equation}
where $\mathcal{M}_t = \{i : x_t^i = [\mathrm{MASK}]\}$ is the set of masked indices at step $t$.

\begin{wraptable}{r}{0.52\linewidth}
\vspace{-10pt}
\centering

\small
\setlength{\tabcolsep}{4.5pt}
\renewcommand{\arraystretch}{1.05}

\begin{tabular}{lccc}
\toprule
Section & Value & Scaffold & Blocks \\
\midrule
\texttt{critical\_objects}      & 12  & 80  & 1 \\
\texttt{explanation}            & 192 & 6   & 6 \\
\texttt{future\_meta\_behavior} & 6   & 18  & 1 \\
\texttt{trajectory}             & 70  & 20  & 3 \\
\midrule
\textbf{Total}                  & \textbf{280} & \textbf{124} & \textbf{11} \\
\bottomrule
\end{tabular}

% \vspace{4pt}

% {\tiny
% \begin{minipage}{0.95\linewidth}
% \begin{lstlisting}
% {
%  "critical_objects": {
%    "nearby_vehicle": "yes", "pedestrian": "no",
%    "cyclist": "no", "traffic_element": "yes",
%    "road_hazard": "no", "weather_condition": "no",
%    "construction": "no", "emergency_vehicle": "no",
%    "animal": "no", "special_vehicle": "no",
%    "conflicting_vehicle": "no",
%    "door_opening_vehicle": "no"
%  },
%  "explanation": "This is an example.",
%  "future_meta_behavior": {
%    "longitudinal": "decelerate",
%    "lateral":      "keep_lane"
%  },
%  "trajectory":
%    "[[ +003.30,-000.01], [ +006.71,-000.04],
%      [ +010.12,-000.09], [ +013.42,-000.16],
%      [ +016.30,-000.24]]"
% }
% \end{lstlisting}
% \end{minipage}
% }

{\tiny
\begin{minipage}{0.95\linewidth}
\begin{lstlisting}
{
  "critical_objects": {
    "nearby_vehicle":"|\val{yes}|", "pedestrian":"|\val{no}|",
    "cyclist":"|\val{no}|", "traffic_element":"|\val{yes}|",
    "road_hazard":"|\val{no}|", "weather_condition":"|\val{no}|",
    "construction":"|\val{no}|", "emergency_vehicle":"|\val{no}|",
    "animal":"|\val{no}|", "special_vehicle":"|\val{no}|",
    "conflicting_vehicle":"|\val{no}|",
    "door_opening_vehicle":"|\val{no}|"
  },
  "explanation": "|\val{This is an example.}|",
  "future_meta_behavior": {
    "longitudinal": "|\val{decelerate}|",
    "lateral": "|\val{keep lane}|"
  },
  "trajectory":
    "|\val{[[ +003.30,-000.01], [ +006.71,-000.04]}|,
      |\val{[ +010.12,-000.09], [ +013.42,-000.16]}|,
      |\val{[ +016.30,-000.24]]}|"
}
\end{lstlisting}
\end{minipage}
}

% \vspace{2pt}

\caption{\textbf{Top}: per-section value/scaffold token counts in our schema. The scaffold accounts for ${\sim}30\%$ of decoded
tokens. \textbf{Bottom}: an example structured output of \modelname. Only \colorbox{valbg}{\texttt{value tokens}} need to be decoded.}
\label{tab:section_tokens}
\vspace{-30pt}
\end{wraptable}

\paragraph{Block-Causal Diffusion.}
Full-sequence bidirectional diffusion~\citep{nie2025large} precludes KV-cache reuse and requires full recomputation at every denoising step. Block Diffusion~\citep{arriola2025block} addresses this by partitioning the output into $B$ blocks of size $d$: $\mathbf{x}_0 = [\mathbf{b}_1, \ldots, \mathbf{b}_B]$, where blocks are generated left-to-right with \emph{bidirectional} attention within each block and \emph{causal} attention across blocks. Formally, block $\mathbf{b}_j$ attends to the full prompt $\mathbf{c}$ and all preceding blocks $\mathbf{b}_{1:j-1}$ (whose KV cache can be reused), but not to future blocks $\mathbf{b}_{j+1:B}$. This recovers KV-cache compatibility while retaining parallel generation within each block.

Fast-dVLM~\citep{wu2025fastdvlm} extends block diffusion to vision-language models via direct conversion from autoregressive VLMs, and introduces \emph{self-speculative decoding}~\citep{wu2025fast_dllm}: for each block, the MDM head drafts all tokens in parallel via bidirectional attention, then an AR head with causal attention verifies the draft sequentially, accepting tokens until the first mismatch plus one bonus token. This achieves significant speedup with quality equivalent to pure AR decoding.

\subsection{Structure-Aware Scaffold Diffusion} \label{sec:scaffold}

\begin{figure}[t]
    \centering
%     \fbox{
%   \begin{minipage}[c][7cm][c]{0.9\linewidth}
%     \centering
%     \textbf{[Pipeline Figure Placeholder]}\\
%     \small (to be replaced)
%   \end{minipage}
% }
\includegraphics[width=\linewidth]{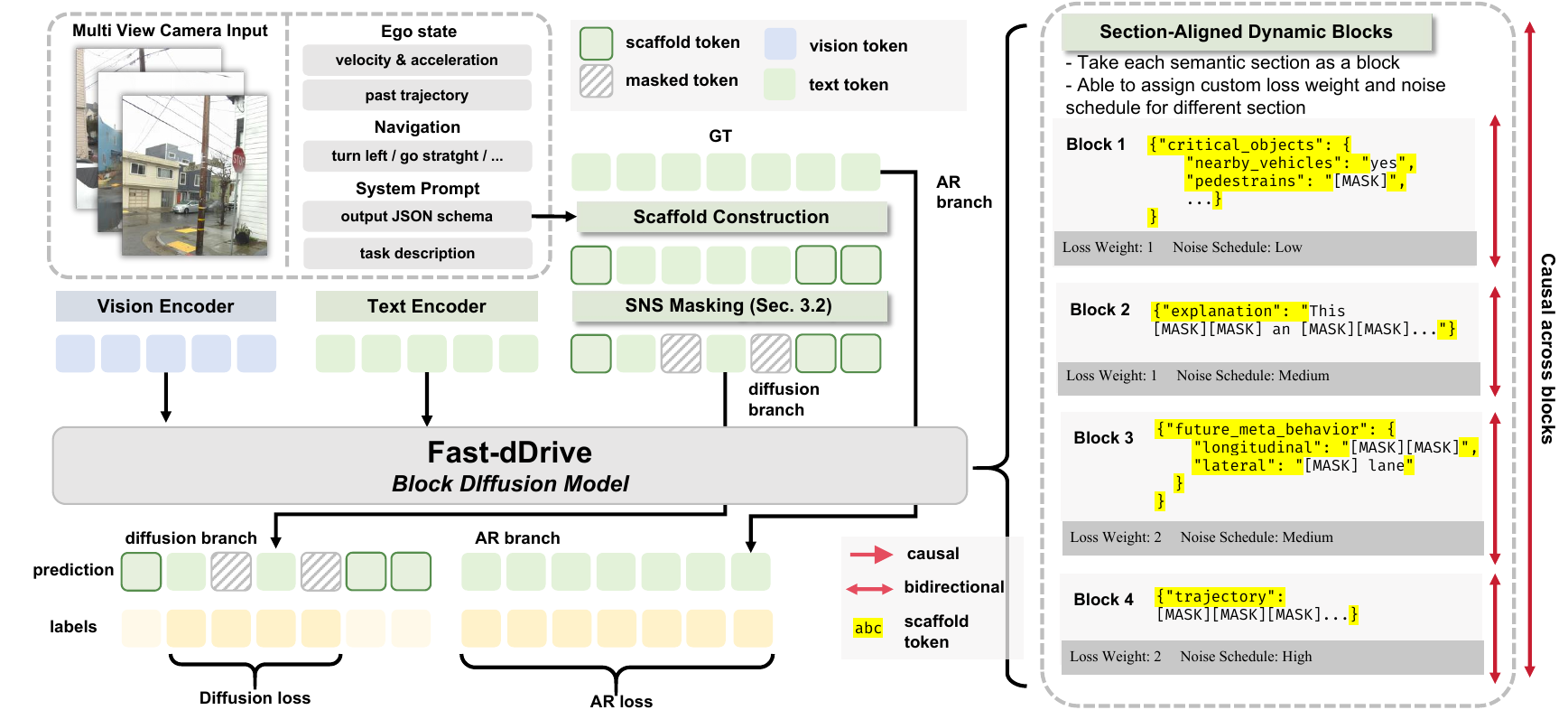}
    \caption{Pipeline of \modelname during training. The structured JSON output is decomposed into four semantic sections; template tokens form a frozen \emph{scaffold} (grey, pre-filled and never masked), while value tokens are denoised in section-aligned blocks with bidirectional attention within each block and strict causal ordering across sections. SASD adds per-section loss weights and Beta noise schedules at training time only.}
    \label{fig:pipeline}
\vspace{-8pt}
\end{figure}

\paragraph{Scaffold Construction and Section-Aligned Blocks.}
Following prior work~\citep{ma2025dvlm,rowe2025poutine,gu2026structuredcot}, the model outputs a structured JSON with four semantic sections: \texttt{critical\_objects} (12 binary detections), \texttt{explanation} (free-form reasoning), \texttt{future\_meta\_behavior} (categorical actions), and \texttt{trajectory} (5 waypoint coordinates over 5\,s). These sections differ dramatically in token count, difficulty, and safety impact.

We exploit the fixed JSON schema by pre-filling all structural tokens (keys, brackets, punctuation) as a frozen \textbf{scaffold} $\hat{\mathbf{x}}_T$, leaving only \emph{value tokens} masked. Let $\mathcal{A}$ denote scaffold (anchor) positions and $\mathcal{E} = \{1{:}L\} \setminus \mathcal{A}$ the editable value positions; the diffusion process operates exclusively on $\mathcal{E}$:
\begin{equation}
\label{eq:scaffold_loss}
\mathcal{L}_{\mathrm{scaffold}}(\theta) = \mathbb{E}_{t, \mathbf{x}_0, \mathbf{x}_t} \left[ -\frac{1}{|\mathcal{M}_t|} \sum_{i \in \mathcal{M}_t} \log p_\theta(x_0^i \mid \mathbf{x}_t, \mathbf{c}) \right], \quad \mathcal{M}_t \subseteq \mathcal{E}.
\end{equation}
This guarantees $100\%$ structural correctness and reduces the denoising workload by ${\sim}30\%$ (Table~\ref{tab:section_tokens}). We further align block boundaries with section boundaries, partitioning each section $s$ into $n_s = \lceil |\mathcal{E}_s| / d \rceil$ blocks. Sections are denoised in the causal order \texttt{CO} $\to$ \texttt{Expl} $\to$ \texttt{FMB} $\to$ \texttt{Traj}, each block providing complete intra-section bidirectional context. Variable-length sections use a \texttt{NULL} token for padding, stripped at inference time.

\paragraph{Safety-Prioritized Training.}
The four sections differ vastly in safety impact: a wrong trajectory coordinate may cause a collision, while a slightly imperfect explanation has no such consequence. We introduce two complementary training-time mechanisms to bias learning capacity toward safety-critical sections.
\emph{Section-weighted loss} assigns each section $s$ a positive scalar weight $w_s$ that scales its per-token cross-entropy:
\begin{equation}
\label{eq:iwl}
\mathcal{L}_{\mathrm{train}}(\theta) = \mathbb{E}_{t, \mathbf{x}_0, \mathbf{x}_t} \left[ -\sum_{s} \frac{w_s}{|\mathcal{M}_t^s|} \sum_{i \in \mathcal{M}_t^s} \log p_\theta(x_0^i \mid \mathbf{x}_t, \mathbf{c}) \right],
\end{equation}
where larger weights are assigned to safety-critical sections so that gradients on hard, high-impact tokens dominate the update.
\emph{Section-adaptive noise} replaces the uniform diffusion schedule with per-section Beta distributions $t_s \sim \mathrm{Beta}(\alpha_s, \beta_s)$, allowing the noise schedule to be tailored to each section's difficulty profile. Concrete values for $\{w_s\}$ and $\{(\alpha_s, \beta_s)\}$ are reported in \S\ref{sec:exp_setup}.
% a more detailed discussion of how the per-section schedule interacts with complementary masking is deferred to Appendix~\ref{app:sasd_noise}.
Both mechanisms incur \textbf{zero inference overhead}.

\paragraph{Joint AR and Diffusion Training.}
Following Fast-dVLM~\citep{wu2025fastdvlm}, we train under a dual-stream objective that combines our section-weighted MDM loss (Eq.~\ref{eq:iwl}) with a token-level causal LM loss $\mathcal{L}_{\mathrm{AR}}$ over the same response labels on the clean stream:
\begin{equation}
\label{eq:joint_loss}
\mathcal{L} = \alpha\, \mathcal{L}_{\mathrm{train}}(\theta) + \beta\, \mathcal{L}_{\mathrm{AR}}(\theta), \quad \alpha = \beta = 0.5.
\end{equation}
The diffusion branch learns parallel value denoising under intra-block bidirectional attention, while the causal branch preserves the pretrained AR decoding capability. As shown in \S\ref{sec:scaffold_spec}, this joint objective is what enables a single trained \modelname{} to expose both a diffusion-only and a self-speculative decoding mode without further fine-tuning.

\subsection{Inference: Section Diffusion and Scaffold Spec} \label{sec:scaffold_spec}

% TODO: Figure — scaffold spec pipeline (draft → verify per block)
\begin{figure}[t]
    \centering
    \includegraphics[width=\linewidth]{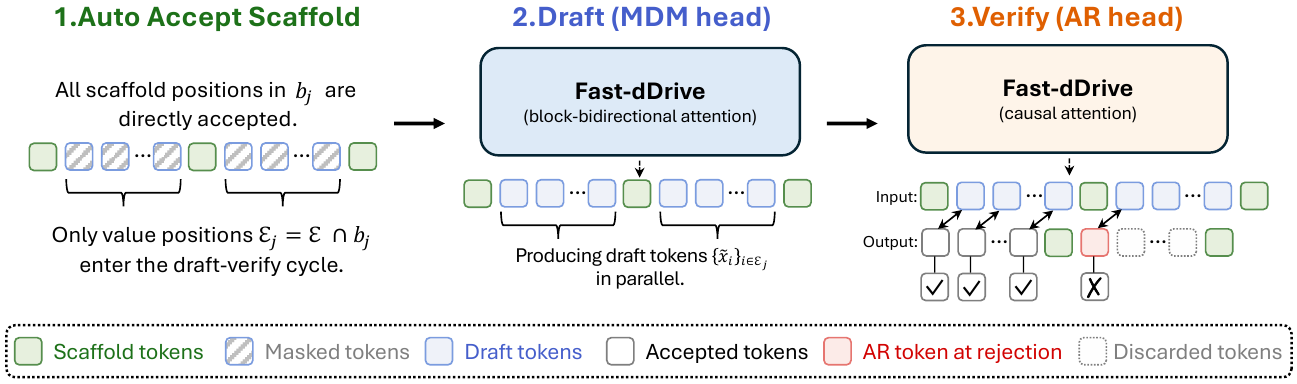}
    \caption{Scaffold Speculative Decoding. For each block: (1) scaffold tokens are auto-accepted; (2) the MDM head drafts value tokens via bidirectional attention; (3) the AR head verifies the draft with causal attention, accepting tokens until the first mismatch plus one bonus token.}
    \label{fig:scaffold_spec}
    \vspace{-10pt}
\end{figure}

Because the joint AR + diffusion objective in Eq.~\eqref{eq:joint_loss} preserves both decoding heads on the same weights, \modelname{} supports two complementary inference modes over the same scaffold and section-aligned blocks, mirroring the dual-mode setup of Fast-dVLM~\citep{wu2025fastdvlm}.

\paragraph{Section Diffusion (SD).}
SD reuses the training-time procedure at inference: starting from the pre-filled scaffold $\hat{\mathbf{x}}_T$, the MDM head iteratively unmasks value positions section by section over the section-aligned dynamic blocks of \S\ref{sec:scaffold}, attending to preceding blocks via cached causal context (i.e., \emph{causal context decoding} in the sense of Fast-dVLM~\citep{wu2025fastdvlm}). KV caches from the scaffold and from earlier sections are reused without recomputation, yielding a diffusion-only baseline that does not invoke the AR head.

\paragraph{Scaffold Speculative Decoding (SS).}
The second mode invokes self-speculative decoding~\citep{wu2025fast_dllm,wu2025fastdvlm}, in which the MDM head drafts a block in parallel and the AR head verifies it sequentially. Vanilla self-spec operates on fixed-size blocks without awareness of scaffolds or section structure; we extend it to \textbf{Scaffold Speculative Decoding} (SS), which exploits the scaffold from \S\ref{sec:scaffold} to further reduce computational overhead while preserving generation quality.

\paragraph{Algorithm.}
Given the pre-filled scaffold $\hat{\mathbf{x}}_T$, Scaffold Spec processes each block $\mathbf{b}_j$ in the section-ordered sequence as follows:
\begin{enumerate}[leftmargin=*,nosep]
    \item \textbf{Auto-accept scaffold}: All scaffold positions within $\mathbf{b}_j$ are directly accepted without drafting or verification. Only value positions $\mathcal{E}_j = \mathcal{E} \cap \mathbf{b}_j$ enter the draft-verify cycle.
    \item \textbf{Draft} (MDM head): A single forward pass with block-bidirectional attention fills all $|\mathcal{E}_j|$ masked value positions simultaneously, producing draft tokens $\{\tilde{x}_i\}_{i \in \mathcal{E}_j}$.
    \item \textbf{Verify} (AR head): A causal forward pass over the entire block computes AR logits. For each value position $i \in \mathcal{E}_j$ in left-to-right order, if $\arg\max p_\theta^{\mathrm{AR}}(\cdot \mid \mathbf{x}_{<i}) = \tilde{x}_i$, the token is accepted; otherwise, the AR token replaces the draft and all subsequent draft tokens are discarded. One bonus token is always accepted at the rejection point.
\end{enumerate}

\paragraph{Efficiency Analysis.}
Each block requires exactly 2 forward passes (draft + verify), regardless of block size. The key speedup over vanilla self-speculative decoding comes from two sources: (1) scaffold tokens are auto-accepted with \emph{zero} forward passes; (2) section-aligned blocks ensure that the MDM draft has complete semantic context, improving draft acceptance rate compared to arbitrary fixed-size blocks. Combined, this yields a remarkable speedup over standard self-speculative decoding.

\subsection{Test-Time Inference Scaling via Shared-Prefix Multi-Trajectory Rollouts} \label{sec:merge}

Scaffold Spec (\S\ref{sec:scaffold_spec}) decodes the structured output deterministically: a single SS pass already returns the model's most-confident trajectory. To convert additional inference compute into additional accuracy, we introduce stochasticity \emph{inside} the AR verifier and average $N$ trajectory rollouts. Two design choices keep this scheme both cheap and quality-preserving.

\begin{wrapfigure}{r}{0.5\linewidth}
    % \vspace{-12pt}
    \centering
    \includegraphics[width=\linewidth]{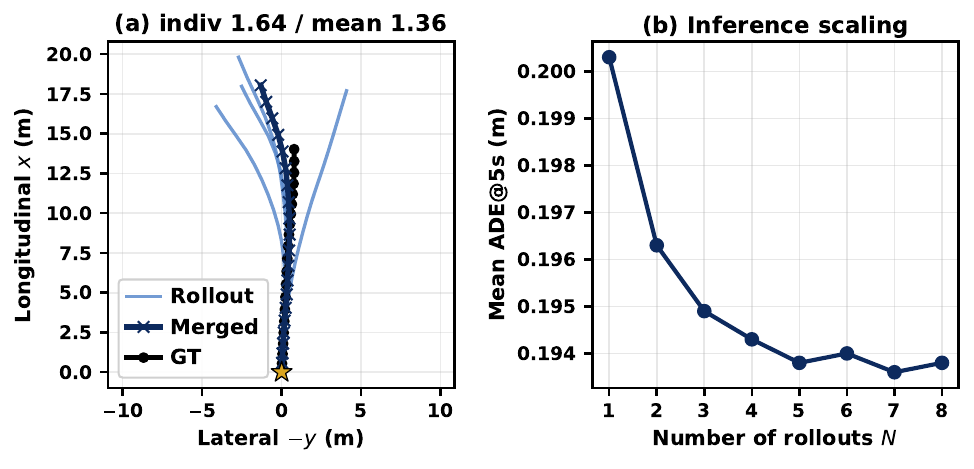}
    \caption{Shared-prefix multi-trajectory rollouts. \textbf{(a)} On a representative WOD-E2E val sample, $N{=}4$ rollouts (light blue) disagree most at the late waypoints, while their mean (dark blue) lies on top of the ground truth (black). \textbf{(b)} Mean ADE@5s on WOD-E2E val decreases monotonically with $N$, confirming the variance-of-the-mean argument.}
    \label{fig:ss_multi_rollout}
    \vspace{-8pt}
\end{wrapfigure}

\textbf{Trajectory-only stochasticity.} The first three sections (\texttt{critical\_objects}, \texttt{explanation}, \texttt{future\_meta\_behavior}) are heavily structured by the schema and have sharply peaked posteriors; sampling them adds no useful diversity and only degrades downstream sections. We therefore keep the AR verifier greedy on the first three sections and only enable softmax sampling once decoding enters the trajectory section.

\textbf{Shared prefix.} Because the first three sections are deterministic, their KV cache is identical across rollouts. We decode them \emph{once}, fork the KV cache $N$ times, and continue Scaffold Spec on the trajectory section $N$ times, each with independent random draws. Since the trajectory section is short relative to the full output, this adds only a fractional cost per extra rollout rather than a full SS pass.

\paragraph{Trajectory averaging.}
Let $\{\boldsymbol{\tau}^{(i)}\}_{i=1}^{N}$ be the $N$ rollout trajectories, each interpolated to 20 waypoints via Jerk-Minimizing Trajectory (JMT) fitting. The output is the equal-weight average
% \begin{equation}
$
\label{eq:merge}
\boldsymbol{\tau}_{\mathrm{out}} \;=\; \frac{1}{N}\sum_{i=1}^{N} \boldsymbol{\tau}^{(i)} .
$
% \end{equation}
By the variance-of-the-mean argument, averaging $N$ rollouts reduces residual variance by a factor of $1/N$ while leaving any deterministic bias unchanged. Each rollout is still produced by Scaffold Spec (only the trajectory-section verifier step is sampled), so per-rollout quality stays close to the deterministic SS baseline, a regime that sampling the verifier on the full output cannot reach. 
% Hyperparameter values, wall-clock measurements, and the full ablation are reported in \S\ref{sec:exp_setup} and Appendix~\ref{app:merge_analysis}.

Figure~\ref{fig:ss_multi_rollout} illustrates the effect: individual rollouts disagree mostly at the late waypoints, their mean lies closer to the ground truth than any single rollout, and mean ADE@5s decreases monotonically with $N$.

\section{Experiments} \label{sec:exp}
\subsection{Experimental Setup} \label{sec:exp_setup}

\paragraph{Datasets.}
We evaluate in open-loop settings on two established benchmarks. \textbf{nuScenes}~\citep{caesar2020nuscenes} contains 1{,}000 urban driving scenes split 700/150/150 for train/val/test, with annotated keyframes sampled at 2\,Hz. \textbf{Waymo Open Dataset End-to-End (WOD-E2E)}~\citep{xu2025wod} comprises 4{,}021 long-tail driving segments of 20\,s each, split 2{,}037/479/1{,}505; for test evaluation, only the first 12\,s of each segment are provided and predictions must be generated from information available up to that point. We adopt the chain-of-thought annotations from dVLM-AD~\citep{ma2025dvlm} for the four-section structured output. Sensor specifications and capture rates are deferred to Appendix~\ref{app:input}.

\paragraph{Input Modalities.}
The model consumes RGB camera frames, ego state, and a high-level navigation command; we use no LiDAR, radar, or HD map. Following prior open-loop VLA work~\citep{ma2025dvlm,rowe2025poutine}, we use 3 past front-camera frames spanning the last 1\,s on nuScenes and the 3 front-facing cameras at the current frame on WOD-E2E; each image is resized so that the longer side is at most $512$\,px before being patchified by Qwen2.5-VL's vision encoder. Frame timestamps, per-view sizing, and the joint-view WOD-E2E variant we explored are detailed in Appendix~\ref{app:input}.

\paragraph{Evaluation Metrics.}
\emph{Planning accuracy.} For nuScenes, following dVLM-AD~\citep{ma2025dvlm}, we report the L2 distance error at 1/2/3\,s horizons. For WOD-E2E, we report Average Displacement Error (ADE) at 3\,s and 5\,s horizons, and Rater Feedback Score (RFS)~\citep{xu2025wod}, a human-aligned trust-region score where higher values indicate better matches to multiple human-rated reference trajectories.
\emph{Inference efficiency.} On a single NVIDIA H100 with batch size 1 we report \emph{Latency} (ms per sample), \emph{TPS} (tokens per second), and \emph{Tok/Step} (tokens committed per model forward pass; AR decoding gives $\text{Tok/Step}{=}1$).

% \begin{table}[t]
% \centering
% \caption{Input settings per dataset and variant. ``Cams.'' lists the cameras used (FL/F/FR = front-left/front/front-right); ``Frame ts.'' gives the relative timestamps of the visual frames in seconds.}
% \label{tab:settings}
% \small
% \begin{tabular}{llccc}
% \toprule
% Dataset & Variant & Cams. & Frame ts.\ (s) & Image size \\
% \midrule
% nuScenes & Multi-frame front (default) & F & $\{-1.0,\, -0.5,\, 0\}$ & longer side $\leq 512$ \\
% \midrule
% WOD-E2E  & 3-view (default)            & FL, F, FR & $\{0\}$ & longer side $\leq 512$ per view \\
% WOD-E2E  & Joint-view (3 views concat) & FL$ \!\mid\! $F$ \!\mid\! $FR & $\{0\}$ & longer side $\leq 512$ (whole panorama) \\
% \bottomrule
% \end{tabular}
% \end{table}

\paragraph{Implementation Details.}
\modelname is built on Qwen2.5-VL-3B~\citep{bai2025qwen2} converted to the Fast-dVLM~\citep{wu2025fastdvlm} block-diffusion architecture and outputs a structured JSON of the four sections defined in \S\ref{sec:scaffold}. We train the model on $8{\times}$H100 GPUs with block-causal attention, fine-tuning for 3 epochs on the WOD-E2E training set. For WOD-E2E we mix the $30$k CoT-annotated samples with an additional $60$k trajectory-only samples (no CoT) to improve trajectory coverage; the nuScenes training set ($23$k samples) is used separately. SASD instantiates Eq.~\eqref{eq:iwl} with section loss weights $\{w_s\} = \{3.0, 2.0, 1.5, 1.0\}$ and Beta noise parameters $\{(\alpha_s, \beta_s)\} = \{(2,1), (1,1.5), (1,2), (1,1)\}$ for \texttt{trajectory}, \texttt{future\_meta\_behavior}, \texttt{critical\_objects}, and \texttt{explanation} respectively. Efficiency benchmarks are measured on a single H100.

\subsection{Main Results} \label{sec:main_results}

\begin{table}[t]
\centering
\caption{Comparison on WOD-E2E \textbf{test set}. ${}^*$: zero-shot (no fine-tuning). ${}^{\dagger}$: measured by us with the original backbone under the same conditions as our model. TPS and Tok/Step are measured on a single H100.}
\label{tab:wod_e2e_test}
\resizebox{0.9\linewidth}{!}{
\begin{tabular}{llccccc}
\toprule
Method & Paradigm & RFS $\uparrow$ & ADE (5s) $\downarrow$ & ADE (3s) $\downarrow$ & TPS $\uparrow$ & Tok/Step $\uparrow$ \\
\midrule
OpenEMMA${}^*$ \citep{xing2025openemma} & AR & 5.158 & 12.476 & 6.684 & -- & 1 \\
LightEMMA${}^*$ \citep{qiao2025lightemma} & AR & 6.517 & 3.740 & 1.705 & -- & 1 \\
NaiveEMMA & AR & 7.528 & 3.018 & 1.320 & -- & 1 \\
AutoVLA \citep{zhouautovla} & AR & 7.557 & 2.958 & 1.351 & \ \ \ \  51.2$^\dagger$ & 1 \\
Poutine-Base \citep{rowe2025poutine} & AR & \textbf{7.909} & 2.940 & 1.270 & \ \ \ \ 51.2$^\dagger$ & 1 \\
\midrule
dVLM-AD \citep{ma2025dvlm} & Diffusion & 7.633 & 3.022 & 1.285 & \ \ 35.2 & 2.82 \\
\midrule
\modelname (Scaffold Spec) & Block Diff. & 7.823  & \textit{2.907}  & \textit{1.254}  & \textbf{210.4} & \textbf{4.90} \\
\quad + Inference scaling (N=4) & Block Diff. & \textit{7.827} & \textbf{2.821} & \textbf{1.240} & \textit{114.7} & 2.76 \\
\bottomrule
\end{tabular}
}
\vspace{-10pt}
\end{table}

\paragraph{WOD-E2E Results.}

\begin{wraptable}{r}{0.44\textwidth}
\vspace{-20pt}
\centering
\caption{L2 Error on nuScenes \textbf{val set}. ${}^*$: zero-shot.}
\vspace{-7pt}
\label{tab:nuscenes}
\small
\resizebox{\linewidth}{!}{
\begin{tabular}{lcccc}
\toprule
\multirow{2}{*}{Method} & \multicolumn{4}{c}{L2 Error (m) $\downarrow$} \\
\cmidrule(lr){2-5}
 & 1s & 2s & 3s & \cellcolor[HTML]{E8E8FF}\textbf{Avg.} \\
\midrule
\multicolumn{5}{l}{\textit{Training-based Policy}} \\
UniAD~\citep{hu2023planning} & 0.20 & 0.42 & 0.75 & \cellcolor[HTML]{E8E8FF}0.46 \\
VAD-Base~\citep{jiang2023vad} & 0.17 & 0.34 & 0.60 & \cellcolor[HTML]{E8E8FF}0.37 \\
BEV-Planner~\citep{li2024ego} & 0.16 & 0.32 & 0.57 & \cellcolor[HTML]{E8E8FF}0.35 \\
\midrule
\multicolumn{5}{l}{\textit{VLMs / VLAs with Reasoning}} \\
OpenEMMA${}^*$~\citep{xing2025openemma} & 1.45 & 3.21 & 3.76 & \cellcolor[HTML]{E8E8FF}2.81 \\
DriveVLM~\citep{tian2024drivevlm} & 0.18 & 0.34 & 0.68 & \cellcolor[HTML]{E8E8FF}0.40 \\
AutoVLA~\citep{zhouautovla} & 0.25 & 0.46 & 0.73 & \cellcolor[HTML]{E8E8FF}0.48 \\
dVLM-AD~\citep{ma2025dvlm} & 0.15 & 0.40 & 0.68 & \cellcolor[HTML]{E8E8FF}0.41 \\
\midrule
\modelname (ours) & \textbf{0.12}  & \textbf{0.33} & \textbf{0.50} & \textbf{0.32}\cellcolor[HTML]{E8E8FF} \\
\bottomrule
\end{tabular}
}

\vspace{-5pt}
\end{wraptable}

Table~\ref{tab:wod_e2e_test} reports planning accuracy and decoding throughput on the WOD-E2E test set against representative AR baselines and the diffusion baseline dVLM-AD. With a single inference run, \modelname (Scaffold Spec) attains the lowest ADE@3s and ADE@5s among the compared methods, and an RFS that surpasses the diffusion baseline dVLM-AD by a clear margin and is competitive with the strongest AR baseline despite our model using neither GRPO post-training nor a larger trajectory pool. Adding the shared-prefix multi-trajectory rollout of \S\ref{sec:merge} further reduces both ADE values at sub-$2{\times}$ wall-clock cost relative to a single Scaffold-Spec pass, since only the trajectory section is rolled out $N$ times from a forked KV cache. On efficiency, \modelname runs at $4{\times}$--$6{\times}$ the decoding throughput of dVLM-AD and the AR baselines while committing ${\sim}5$ tokens per model forward pass, giving a clear advantage on the accuracy--efficiency Pareto frontier.

\paragraph{nuScenes Results.}

Table~\ref{tab:nuscenes} reports L2 errors on the nuScenes validation set following the dVLM-AD~\citep{ma2025dvlm} protocol. \modelname achieves the lowest average L2 among the listed VLM/VLA systems with reasoning, with consistent gains over the diffusion and AR-with-CoT baselines across all three horizons; it also matches or improves upon classical training-based driving policies that lack interpretable reasoning. Combined with the WOD-E2E results, this indicates that our structure-aware design transfers between predominantly nominal urban driving and long-tail scenarios without per-dataset tuning.

\subsection{Efficiency \& Performance Analysis} \label{sec:efficiency}

Table~\ref{tab:efficiency} compares \modelname inference variants against the AR baseline (Qwen2.5-VL-3B trained with the same data and recipe but standard autoregressive decoding) and dVLM-AD on the WOD-E2E validation set (single H100, batch size 1).

Among the \modelname variants, Scaffold Spec achieves the lowest latency and highest throughput, nearly doubling the TPS of vanilla self-speculative decoding while matching its accuracy. The speedup stems from scaffold auto-acceptance, which removes ${\sim}30\%$ of tokens from the draft-verify loop. Compared to dVLM-AD, which requires full-sequence recomputation at every denoising step, Scaffold Spec achieves roughly $6{\times}$ the throughput by combining block-level KV-cache reuse with scaffold-aware speculative acceptance. The AR baseline, despite sharing the same backbone and training data, is both slower (limited to one token per forward pass) and less accurate than the block-diffusion variants, suggesting that bidirectional context within blocks produces more globally consistent trajectories than purely sequential decoding. Section Diffusion yields competitive throughput but slightly higher ADE than the speculative variants, indicating that causal AR verification contributes meaningfully to trajectory quality; this accuracy gap motivates the shared-prefix rollout scheme of \S\ref{sec:merge}. All \modelname variants substantially outperform dVLM-AD on both ADE and RFS, confirming that section-aware block diffusion with SASD training is more effective than full-sequence diffusion for structured driving outputs. Finally, integrating Scaffold Spec into SGLang~\citep{zheng2024sglang} yields an additional ${\sim}3{\times}$ speedup via optimized kernels and CUDA graph, demonstrating that the algorithmic gains compose well with system-level optimizations.

\begin{table}[hbtp]
\centering
\vspace{-3mm}
\caption{Inference efficiency and accuracy comparison on WOD-E2E val set. Latency: average wall-clock time per sample. TPS: tokens per second (including scaffold tokens). Tok/Step: effective tokens committed per model forward pass.}
\label{tab:efficiency}
\resizebox{1.0\linewidth}{!}{
\begin{tabular}{llcccccc}
\toprule
Method & Decoding & Latency (ms) $\downarrow$ & TPS $\uparrow$ & Tok/Step $\uparrow$ & ADE (3s) $\downarrow$ & ADE (5s) $\downarrow$ & RFS $\uparrow$ \\
\midrule
AR Baseline (Qwen2.5-VL-3B) & Autoregressive & 7855 & 51.6 & 1 & 0.839 & 2.083 & 7.931 \\
dVLM-AD (Full-seq MDM) & Iterative Denoise & 9575 {\color{red}(0.8$\times$)} & 35.2 & 2.82 & 1.119 & 3.024 & 7.187 \\
\midrule
\modelname (Self-Spec) & Draft + Verify & 3714 {\color{teal}(2.1$\times$)} & 109.0 & 2.41 & 0.811 & 1.973 & 7.959 \\
\modelname (Section Diffusion) & Iterative MDM & 3006 {\color{teal}(2.6$\times$)} & 134.4 & 3.28 & 0.840 & 2.058 & 7.928  \\
\quad +Scaffold Spec & Scaffold + D\&V & 1919 {\color{teal}(4.1$\times$)} & 210.4 & 4.90 & 0.812 & 1.982 & 7.934 \\
\quad \quad +SGLang & Scaffold + D\&V & \ \ 665 {\color{teal}(11.8$\times$)} & 608.5 & 4.93 & 0.816 & 1.995 & 7.914 \\

\bottomrule
\end{tabular}
}
\vspace{-3mm}
\end{table}

\subsection{Ablation Studies} \label{sec:ablation}

% \paragraph{Effect of SASD Training.}

\begin{wraptable}{r}{0.38\textwidth}
\vspace{-10pt}
\centering
\caption{SASD ablation (WOD-E2E val, Scaffold Spec). IWL: Section-Importance-Weighted Loss. SNS: Section-Adaptive Noise Schedule.}
\label{tab:ablation_sasd}
\small
\begin{tabular}{cccc}
\toprule
IWL & SNS & ADE (5s) $\downarrow$ & RFS $\uparrow$ \\
\midrule
 &  & 2.028 & 7.735 \\
\checkmark &  & 2.003 & 7.855 \\
 & \checkmark & 2.050 & 7.807 \\
\checkmark & \checkmark & 2.034 & \textbf{7.916} \\
\bottomrule
\end{tabular}
\vspace{-10pt}
\end{wraptable}

% \begin{wraptable}{r}{0.5\textwidth}
% \vspace{-10pt}
% \centering
% \caption{SASD ablation (WOD-E2E val, Scaffold Spec inference). IWL: Section-Importance-Weighted Loss. SNS: Section-Adaptive Noise Schedule.}
% \label{tab:ablation_sasd}
% \small
% \begin{tabular}{ccccc}
% \toprule
% IWL & SNS & RFS $\uparrow$ \\
% \midrule
%  &  &  7.735 \\
% \checkmark &   & 7.855 \\
%  & \checkmark  & 7.807 \\
% \checkmark & \checkmark &   7.916 \\
% \bottomrule
% \end{tabular}
% \vspace{-10pt}
% \end{wraptable}

% RFS-ADE_3s-ADE_5s: checkpoint-2700  ansonly:score=[4.93963],RFS=7.80696,ADE_3s=0.817846,ADE_5s=2.04949  iwlonly:score=[5.05482],RFS=7.85536,ADE_3s=0.797139,ADE_5s=2.0034  noans_noiwl:score=[4.90057],RFS=7.73496,ADE_3s=0.806788,ADE_5s=2.02761  mixed_20260425:score=[5.05781],RFS=7.91631,ADE_3s=0.824637,ADE_5s=2.03386  min_margin=[0.0390642] worst_margin=[0.0390642]

We ablate the two components of SASD training (Section-Importance-Weighted Loss, IWL; Section-Adaptive Noise Schedule, SNS) by re-training under each of the four on/off combinations while holding all other factors fixed; results are in Table~\ref{tab:ablation_sasd}. IWL is the primary contributor: by up-weighting trajectory and meta-behavior tokens, it directly amplifies the gradient on the positions most critical for planning quality, yielding a clear RFS improvement over the uniform-weight baseline. SNS alone provides a smaller but complementary gain by biasing the noise schedule toward harder denoising configurations for safety-critical sections.
% giving the model more practice on the token positions that matter most. 
When combined, IWL and SNS achieve the best RFS among all configurations, indicating that loss weighting and noise shaping address complementary aspects of the training objective.

Table~\ref{tab:efficiency} further confirms that Scaffold Spec is the most efficient inference method at no accuracy cost relative to Self-Spec, while Section Diffusion offers a useful alternative when diversity is needed (see \S\ref{sec:merge}). For test-time scaling, Figure~\ref{fig:ss_multi_rollout}(b) shows that ADE@5s decreases monotonically with the number of trajectory rollouts $N$; we adopt $N{=}4$ as the default, which provides a favorable accuracy--latency trade-off.

\section{Conclusion}

We presented \modelname, a block-diffusion VLA that exploits the inherent structure of driving outputs to simultaneously advance planning accuracy and inference efficiency. By treating deterministic schema tokens as a frozen scaffold, aligning diffusion blocks with semantic sections, and prioritizing safety-critical tokens during training, \modelname achieves state-of-the-art trajectory accuracy at $6{\times}$ the throughput of full-sequence diffusion baselines, demonstrating that structured generation and efficient decoding are complementary rather than conflicting objectives. The shared-prefix multi-trajectory rollout scheme further shows that block diffusion naturally admits a low-cost test-time scaling axis unavailable to AR models. We believe these results point toward a broader principle: when model outputs have known structure, encoding that structure into the diffusion process yields compounding gains in both quality and speed.

% We presented Fast-dDrive, a block-diffusion VLA for end-to-end autonomous driving that achieves state-of-the-art planning accuracy while fundamentally redefining inference efficiency for high-capacity agents. Our framework hinges on three synergistic design choices: (1) a frozen scaffold that pre-fills deterministic JSON structures, ensuring 100\% schema validity while pruning the decoding workload by ${\sim}30\%$; (2) section-aligned dynamic blocks paired with Section-Aware Structured Diffusion (SASD), which concentrate model capacity on safety-critical value tokens at zero inference overhead; and (3) Scaffold Speculative Decoding, which leverages the structural scaffold to accelerate AR verification, delivering ${\sim}5$ tokens per forward pass and yielding a $6\times$ throughput increase over full-sequence diffusion baselines.Empirical evaluations on WOD-E2E and nuScenes demonstrate that Fast-dDrive sets a new speed-accuracy frontier, reducing average L2 error by $22\%$ compared to prior diffusion models. Furthermore, our shared-prefix multi-trajectory rollout scheme offers a scalable mechanism to trade marginal compute for enhanced robustness. Crucially, when integrated with the SGLang inference engine, Fast-dDrive delivers a massive $12\times$ throughput speedup over AR baselines (rising to 608 TPS), demonstrating that large-scale VLAs can effectively bridge the gap toward the stringent efficiency demands of real-time on-vehicle deployment without accuracy compromises.

\clearpage
\appendix

\section{Input Modalities: Full Details} \label{app:input}

This appendix gives the complete per-dataset description of the visual inputs and image-processing pipeline summarized in \S\ref{sec:exp_setup}.

\paragraph{nuScenes.}
Following dVLM-AD~\citep{ma2025dvlm}, we use only the front camera (\texttt{CAM\_FRONT}) and provide three frames covering the past $1$\,s at relative timestamps $t \in \{-1.0,\, -0.5,\, 0\}$\,s relative to the prediction time. The three frames are presented to the model in chronological order so that short-horizon ego-motion cues (acceleration, yaw rate, immediate-future intent) are recoverable from the visual input alone, complementing the explicit ego-state vector. We do not use the side or rear cameras of nuScenes; the front-camera-only setup matches the dVLM-AD evaluation protocol and keeps the comparison fair.

\paragraph{WOD-E2E.}
Following Poutine~\citep{rowe2025poutine}, we use the three front-facing cameras (\texttt{FRONT\_LEFT}, \texttt{FRONT}, \texttt{FRONT\_RIGHT}) of WOD-E2E at the current frame ($t = 0$\,s), rather than all eight surround views. Although omitting the side and rear views reduces contextual coverage, in our preliminary experiments we found the three forward views sufficient for the open-loop planning task while keeping the visual token budget tractable for our $3$B-parameter backbone.

\paragraph{Joint-view variant on WOD-E2E.}
As an additional variant, we explore a \emph{joint-view} input in which the three WOD-E2E front views are horizontally concatenated (\texttt{FRONT\_LEFT}$|$\texttt{FRONT}$|$\texttt{FRONT\_RIGHT}) into a single panoramic image before resizing. Compared with feeding the three views as three separate images, the joint-view variant trades per-view resolution for a single-image visual context and roughly halves the visual token count, at the cost of a wider but lower-resolution panorama.

\paragraph{Image resizing.}
For both datasets and both WOD-E2E variants, we resize each input image so that its longer side is at most $512$ pixels, with the aspect ratio preserved, before patchification by Qwen2.5-VL's native vision encoder. We do not perform any cropping, color jittering, or other photometric augmentation at inference time; at training time we follow the standard Qwen2.5-VL preprocessing pipeline.

\paragraph{Non-visual inputs.}
On both datasets, the model is conditioned on the ego state at the prediction time (position, velocity, acceleration, yaw, yaw rate, in the ego frame) and a high-level natural-language navigation command (e.g., ``go straight,'' ``turn left at the next intersection''), serialized into the prompt prior to the visual tokens. We do not use LiDAR, radar, HD maps, or any other auxiliary sensor modality.

\section{Qualitative Case Studies} \label{app:case_study}

We present five qualitative examples drawn from \emph{five different} Waymo end-to-end driving scenes to illustrate \modelname{}'s behaviour across diverse driving regimes: a planned left turn at night, lane-following behind a pickup truck, a right turn into a side street, a green-light cruise through a signalised intersection, and a wet-weather left turn at a stop-controlled intersection (the same scene used in our demo video, \S\ref{sec:exp_setup}). In each figure the left panel is the cylindrical panorama (\texttt{FRONT\_LEFT} $\!|\!$ \texttt{FRONT} $\!|\!$ \texttt{FRONT\_RIGHT}) with the \emph{ground-truth} future trajectory overlaid (blue gradient, near $\rightarrow$ far), and the right panel is the top-down trajectory plot of \modelname{}'s predicted waypoints in vehicle frame (green gradient, near $\rightarrow$ far) — matching the demo-video style. Below each figure we list the raw JSON Chain-of-Thought that \modelname{} produced for that frame (Scaffold-Spec, single rollout); only the \colorbox{valbg}{\texttt{value tokens}} are decoded by the model. The model's predictions match the GT trajectory direction in every case.

\begin{figure}[t]
    \centering
    \includegraphics[width=\linewidth]{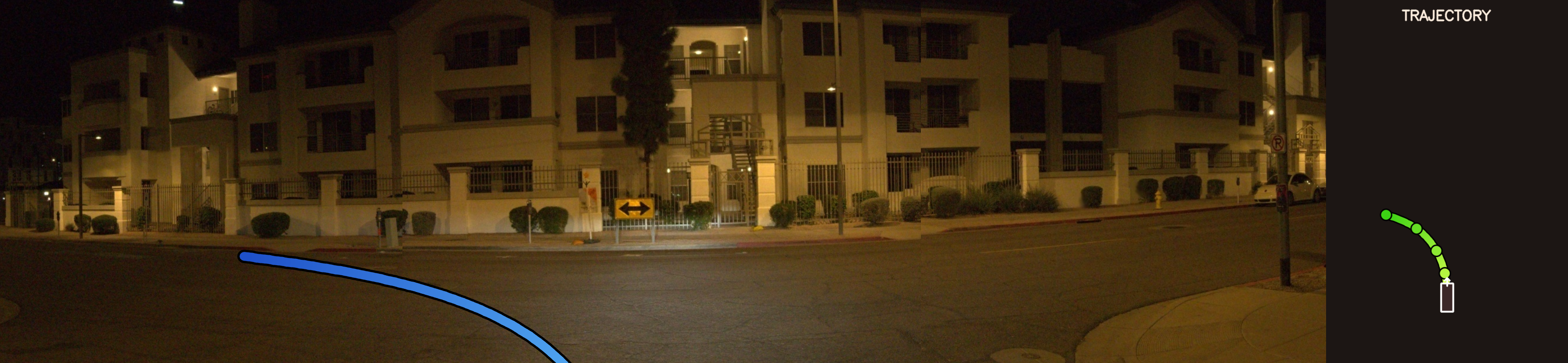}
\vspace{4pt}
{\tiny
\begin{minipage}{0.95\linewidth}
\begin{lstlisting}
{
  "critical_objects": {
    "nearby_vehicle":"|\val{no}|", "pedestrian":"|\val{no}|", "cyclist":"|\val{no}|",
    "construction":"|\val{no}|", "traffic_element":"|\val{no}|", "weather_condition":"|\val{no}|",
    "road_hazard":"|\val{no}|", "emergency_vehicle":"|\val{no}|", "animal":"|\val{no}|",
    "special_vehicle":"|\val{no}|", "conflicting_vehicle":"|\val{no}|", "door_opening_vehicle":"|\val{no}|"
  },
  "future_meta_behavior": { "longitudinal":"|\val{speed up}|", "lateral":"|\val{left turn}|" },
  "explanation":"|\val{There are no visible vehicles, pedestrians, cyclists, construction, or traffic elements in the immediate vicinity that would affect the ego vehicle's path within the next 3 seconds. The road ahead appears clear with no obstacles or hazards requiring the ego vehicle to slow down, yield, or change lanes. The ego vehicle is expected to continue accelerating and make a left turn smoothly without interference.}|",
  "trajectory":"|\val{[[+1.27, +0.03], [+4.21, +0.42], [+8.20, +1.90], [+12.08, +5.38], [+14.50, +10.72]]}|"
}
\end{lstlisting}
\end{minipage}}
    \caption{\textbf{Night left turn into a clear cross street.} With no critical objects in the scene, \modelname{} commits to \texttt{speed up / left turn} and emits a smoothly curving trajectory whose 5-second arc tracks the GT (blue) tightly across the FRONT and FRONT\_LEFT cameras.}
    \label{fig:case_left_turn}
\end{figure}

\begin{figure}[t]
    \centering
    \includegraphics[width=\linewidth]{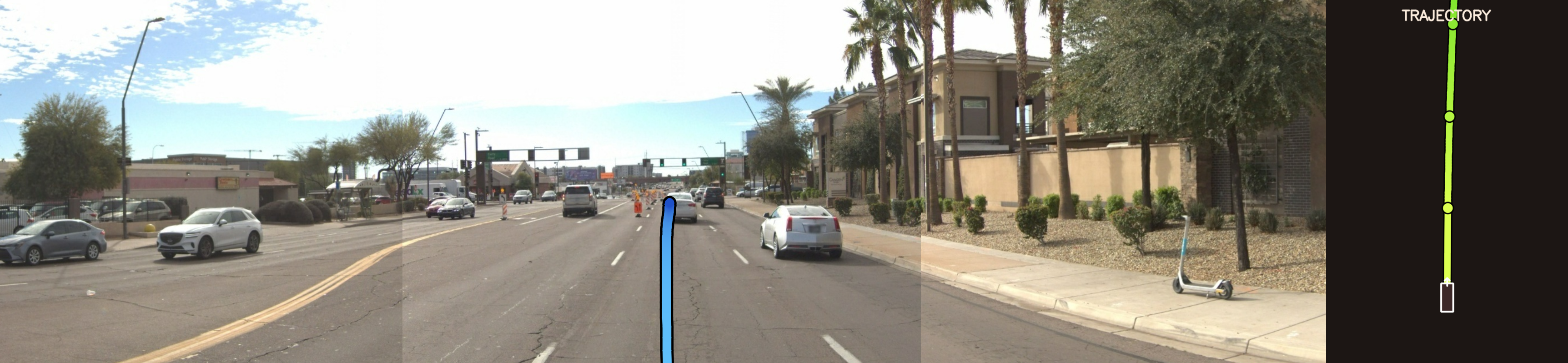}
\vspace{4pt}
{\tiny
\begin{minipage}{0.95\linewidth}
\begin{lstlisting}
{
  "critical_objects": {
    "nearby_vehicle":"|\val{yes}|", "pedestrian":"|\val{no}|", "cyclist":"|\val{no}|",
    "construction":"|\val{no}|", "traffic_element":"|\val{no}|", "weather_condition":"|\val{no}|",
    "road_hazard":"|\val{no}|", "emergency_vehicle":"|\val{no}|", "animal":"|\val{no}|",
    "special_vehicle":"|\val{no}|", "conflicting_vehicle":"|\val{no}|", "door_opening_vehicle":"|\val{no}|"
  },
  "future_meta_behavior": { "longitudinal":"|\val{keep speed}|", "lateral":"|\val{lane follow}|" },
  "explanation":"|\val{There is a nearby vehicle visible ahead in the front camera, approximately in the ego lane, which may influence the ego vehicle's speed and lane keeping. This vehicle's presence requires the ego vehicle to maintain a safe following distance and possibly adjust speed. No pedestrians, cyclists, construction, or traffic elements are present that would affect the ego vehicle's immediate path. No conflicting vehicles or other hazards are detected that would require yielding or lane changes.}|",
  "trajectory":"|\val{[[+15.66, -0.10], [+31.65, -0.42], [+47.85, -1.02], [+64.15, -1.85], [+80.46, -2.77]]}|"
}
\end{lstlisting}
\end{minipage}}
    \caption{\textbf{Lane follow with a pickup truck ahead.} The model flags the single lead vehicle as critical, picks \texttt{lane follow / keep speed}, and emits a long forward trajectory that maintains the lane while keeping pace with the lead. GT (blue) and the predicted final waypoint agree on the longitudinal range.}
    \label{fig:case_pickup_truck}
\end{figure}

\begin{figure}[t]
    \centering
    \includegraphics[width=\linewidth]{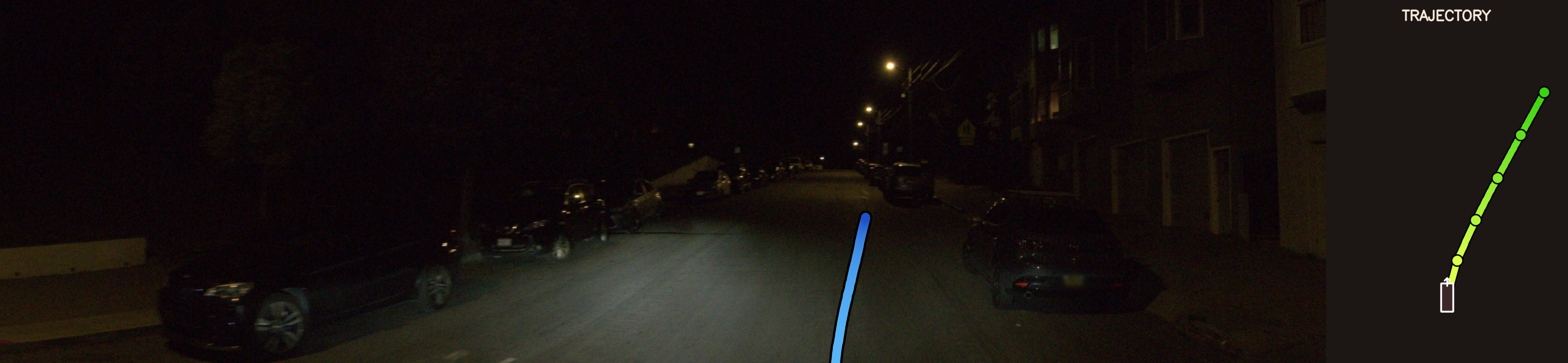}
\vspace{4pt}
{\tiny
\begin{minipage}{0.95\linewidth}
\begin{lstlisting}
{
  "critical_objects": {
    "nearby_vehicle":"|\val{yes}|", "pedestrian":"|\val{no}|", "cyclist":"|\val{no}|",
    "construction":"|\val{no}|", "traffic_element":"|\val{no}|", "weather_condition":"|\val{no}|",
    "road_hazard":"|\val{no}|", "emergency_vehicle":"|\val{no}|", "animal":"|\val{no}|",
    "special_vehicle":"|\val{no}|", "conflicting_vehicle":"|\val{no}|", "door_opening_vehicle":"|\val{no}|"
  },
  "future_meta_behavior": { "longitudinal":"|\val{keep speed}|", "lateral":"|\val{right turn}|" },
  "explanation":"|\val{There is a nearby vehicle visible ahead in the front camera, positioned in the ego lane, which may influence the ego vehicle's speed and lateral positioning. This vehicle's presence requires the ego vehicle to adjust longitudinal behavior, likely slowing or maintaining speed to keep a safe distance. No pedestrians, cyclists, or conflicting vehicles are observed that would require yielding or stopping. No traffic elements or road hazards are present to affect immediate behavior. The ego vehicle is expected to execute a right lateral maneuver based on the trajectory, aligning with the right turn, while adjusting speed accordingly.}|",
  "trajectory":"|\val{[[+6.50, -1.77], [+13.50, -5.00], [+20.88, -8.80], [+28.40, -12.88], [+35.90, -17.00]]}|"
}
\end{lstlisting}
\end{minipage}}
    \caption{\textbf{Sharp right turn off a multi-lane road.} \modelname{} flips the lateral channel to \texttt{right turn} and emits a deeply curving trajectory whose lateral excursion reaches $-17$\,m at 5\,s. The predicted arc tracks the GT (blue) closely across the FRONT $\rightarrow$ FRONT\_RIGHT seam.}
    \label{fig:case_right_turn}
\end{figure}

\begin{figure}[t]
    \centering
    \includegraphics[width=\linewidth]{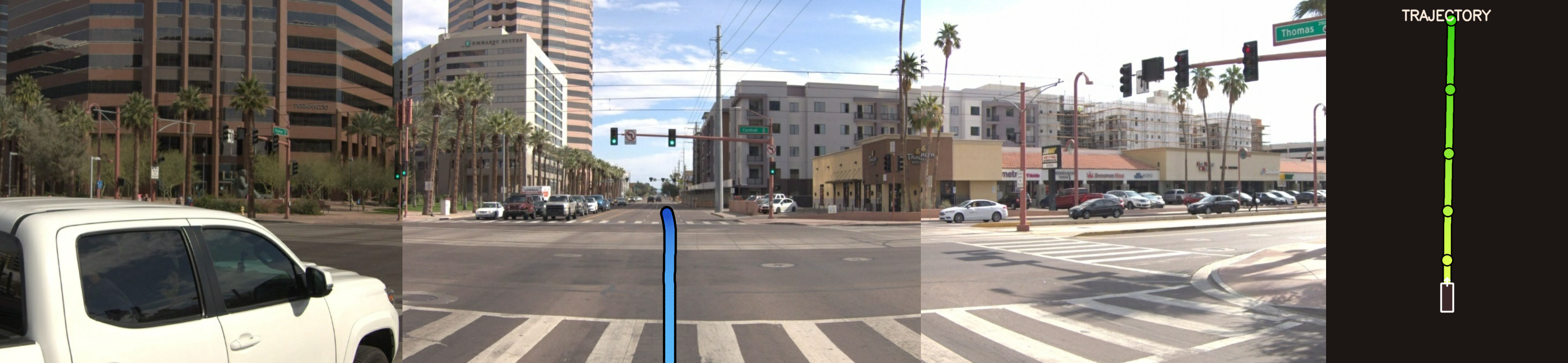}
\vspace{4pt}
{\tiny
\begin{minipage}{0.95\linewidth}
\begin{lstlisting}
{
  "critical_objects": {
    "nearby_vehicle":"|\val{no}|", "pedestrian":"|\val{no}|", "cyclist":"|\val{no}|",
    "construction":"|\val{no}|", "traffic_element":"|\val{yes}|", "weather_condition":"|\val{no}|",
    "road_hazard":"|\val{no}|", "emergency_vehicle":"|\val{no}|", "animal":"|\val{no}|",
    "special_vehicle":"|\val{no}|", "conflicting_vehicle":"|\val{no}|", "door_opening_vehicle":"|\val{no}|"
  },
  "future_meta_behavior": { "longitudinal":"|\val{speed up}|", "lateral":"|\val{go straight}|" },
  "explanation":"|\val{The traffic lights visible ahead in the front camera are relevant as they regulate the ego vehicle's movement through the intersection. They are currently green, allowing the ego vehicle to proceed. No nearby vehicles, pedestrians, or other hazards are present that would require yielding or stopping. The ego vehicle is expected to continue moving forward while monitoring the traffic signal for any changes, adjusting speed accordingly to maintain safe and lawful travel through the intersection.}|",
  "trajectory":"|\val{[[+6.40, -0.02], [+14.62, -0.15], [+23.52, -0.62], [+32.12, -1.76], [+39.99, -3.47]]}|"
}
\end{lstlisting}
\end{minipage}}
    \caption{\textbf{Green-light cruise through a signalised intersection.} The model correctly reads the green signal, tags only \texttt{traffic\_element} as critical, and commits to \texttt{speed up / go straight}. The forward trajectory matches the GT (blue) heading down the same lane through the junction.}
    \label{fig:case_traffic_light}
\end{figure}

\begin{figure}[t]
    \centering
    \includegraphics[width=\linewidth]{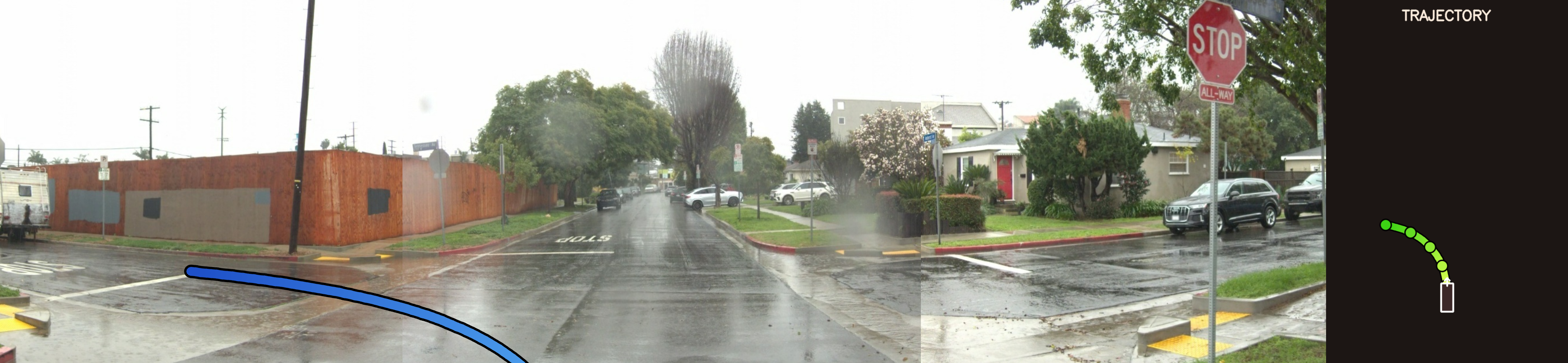}
\vspace{4pt}
{\tiny
\begin{minipage}{0.95\linewidth}
\begin{lstlisting}
{
  "critical_objects": {
    "nearby_vehicle":"|\val{no}|", "pedestrian":"|\val{no}|", "cyclist":"|\val{no}|",
    "construction":"|\val{no}|", "traffic_element":"|\val{yes}|", "weather_condition":"|\val{yes}|",
    "road_hazard":"|\val{no}|", "emergency_vehicle":"|\val{no}|", "animal":"|\val{no}|",
    "special_vehicle":"|\val{no}|", "conflicting_vehicle":"|\val{no}|", "door_opening_vehicle":"|\val{no}|"
  },
  "future_meta_behavior": { "longitudinal":"|\val{slow down}|", "lateral":"|\val{left turn}|" },
  "explanation":"|\val{There is a stop sign clearly visible on the right side of the intersection, which is a critical traffic element requiring the ego vehicle to yield or stop before proceeding. Additionally, the road surface appears wet from rain, indicating slippery conditions that may affect braking distance and vehicle control. No nearby vehicles, pedestrians, or other conflicting objects are present that would immediately impact the ego vehicle's path. The ego vehicle must account for the stop sign and weather conditions to safely navigate the intersection, potentially slowing down and ensuring safe stopping or yielding before continuing the planned left turn.}|",
  "trajectory":"|\val{[[+2.12, +0.09], [+5.50, +0.88], [+8.85, +3.00], [+11.38, +6.40], [+12.73, +10.77]]}|"
}
\end{lstlisting}
\end{minipage}}
    \caption{\textbf{Wet-weather left turn at a stop-controlled intersection} (same scene as the demo video). \modelname{} jointly tags the stop sign (\texttt{traffic\_element}) and the wet road (\texttt{weather\_condition}), commits to \texttt{slow down / left turn}, and emits a sharply curving trajectory that swings from FRONT into FRONT\_LEFT. The cylindrical projection (\S\ref{app:input}) keeps the arc geometrically continuous across the camera seam, and the predicted curve overlays the GT (blue) closely for the entire 5\,s.}
    \label{fig:case_rain_left_turn}
\end{figure}

These five cases together span the most informative planning regimes encountered in our evaluation: a deliberate steering manoeuvre with a clear road, longitudinal-only following behind another vehicle, a sharp lateral commit at a turn, a green-light cruise that exercises the traffic-element detection head, and a wet-weather left turn that demands joint reasoning over traffic infrastructure and adverse road conditions. In every case the structured Chain-of-Thought emitted by Scaffold-Spec is concise, self-consistent with the visible scene, and the predicted trajectory follows the GT direction.

\section{Limitations} \label{app:limit}

While Fast-dDrive significantly improves the throughput and planning accuracy of driving VLAs, it has certain limitations. First, our Scaffold Construction relies on a fixed JSON schema; while this covers the majority of current end-to-end driving tasks, it may require manual template adjustment if the task definition (e.g., the number of detected objects or the granularity of reasoning) changes fundamentally. Second, the shared-prefix inference scaling provides accuracy gains at the cost of some additional compute; while this cost is fractional, it may still be constrained in extremely low-latency edge environments where even a single additional forward pass is prohibitive. Finally, our current evaluation is focused on open-loop benchmarks. Although these are standard for assessing planning quality against human experts, future work should involve closed-loop simulations to further validate the model's reactive capabilities in dynamic environments.

\clearpage
\bibliography{references}
\bibliographystyle{colm2026_conference}

\end{document}